\documentclass[letterpaper, 10 pt, conference]{ieeeconf}  

\IEEEoverridecommandlockouts                              

\usepackage{times}
\usepackage{graphicx}
\DeclareGraphicsExtensions{.pdf,.png,.jpg}
\usepackage{amsmath}
\DeclareMathOperator*{\argmax}{argmax} 
\DeclareMathOperator*{\argmin}{argmin} 
\usepackage{amssymb}  
\usepackage{makecell}
\usepackage{booktabs}
\usepackage{multirow}
\usepackage{here}
\usepackage{kotex}
\usepackage{soul, color}

\title{\LARGE \bf
	Anytime 3D Object Reconstruction using Multi-modal Variational Autoencoder
}

\author{Hyeonwoo Yu and Jean Oh
 	\thanks{This work was funded in part by the AI-Assisted Detection and Threat Recognition Program through US ARMY ACC-APG-RTP (Contract No. W911NF1820218), ``Leveraging Advanced Algorithms, Autonomy, and Artificial Intelligence (A4I) to enhance National Security and Defense'' and Air Force Office of Scientific Research (Award No. FA2386-17-1-4660), and was supported by Institute of Information \& communications Technology Planning \& Evaluation(IITP) grant funded by the Korea government(MSIT)
(No.2020-0-01336, Artificial Intelligence Graduate School Program(UNIST))}
 	\thanks{Hyeonwoo Yu is with the School of Electrical and Computer Engineering, Ulsan National Institute of Science and Technology (UNIST), Ulsan, South Korea
	{\tt\small hwyu2019@gmail.com}}%
	\thanks{Jean Oh is affiliated with the Robotics Institute of Carnegie Mellon University, Pittsburgh, PA 15213, USA
	{\tt\small hyaejino@andrew.cmu.edu}}%
	
}

\begin{document}
	
\maketitle
\thispagestyle{empty}
\pagestyle{empty}

\begin{abstract}
	For effective human-robot teaming, it is important for the robots to be able to share their visual perception with the human operators.
	In a harsh remote collaboration setting, data compression techniques such as autoencoder can be utilized to obtain and transmit the data in terms of latent variables in a compact form.
	In addition, to ensure real-time runtime performance even under unstable environments, an anytime estimation approach is desired that can reconstruct the full contents from incomplete information.
	In this context, we propose a method for imputation of latent variables whose elements are partially lost.
	To achieve the anytime property with only a few dimensions of variables, exploiting prior information of the category-level is essential.
	A prior distribution used in variational autoencoders is simply assumed to be isotropic Gaussian regardless of the labels of each training datapoint. This type of flattened prior makes it difficult to perform imputation from the category-level distributions.
	We overcome this limitation by exploiting a category-specific multi-modal prior distribution in the latent space.
	The missing elements of the partially transferred data can be sampled, by finding a specific modal according to the remaining elements.
	Since the method is designed to use partial elements for anytime estimation, it can also be applied for data over-compression.
	Based on the experiments on the ModelNet and Pascal3D datasets, the proposed approach shows consistently superior performance over autoencoder and variational autoencoder up to 70\% data loss. 
\end{abstract}

\section{Introduction}
	When a human operator is teaming with robots in a remote location, establishing a shared visual perception of the remote location is crucial for a successful team operation.  
	For reliable scene understanding, object recognition is a key element for semantic scene reconstruction and object-oriented simultaneous localization and mapping (SLAM)~\cite{slam++,categorySpecificSLAM,yang2019cubeslam,bowman2017probabilistic,doherty2019multimodal}.
	In this case, 3D shape understanding of object can be exploited for semantic reconstruction~\cite{renderforcnn, dataDriven3Dvoxel,fu2021single}.
	This 3D information can be obtained by directly scanning objects using LIDAR or depth camera, or by network estimation from other visual sensory data such as 2D images.
	
	
	\begin{figure}[t]%
		\centering%
		\includegraphics[scale=0.055]{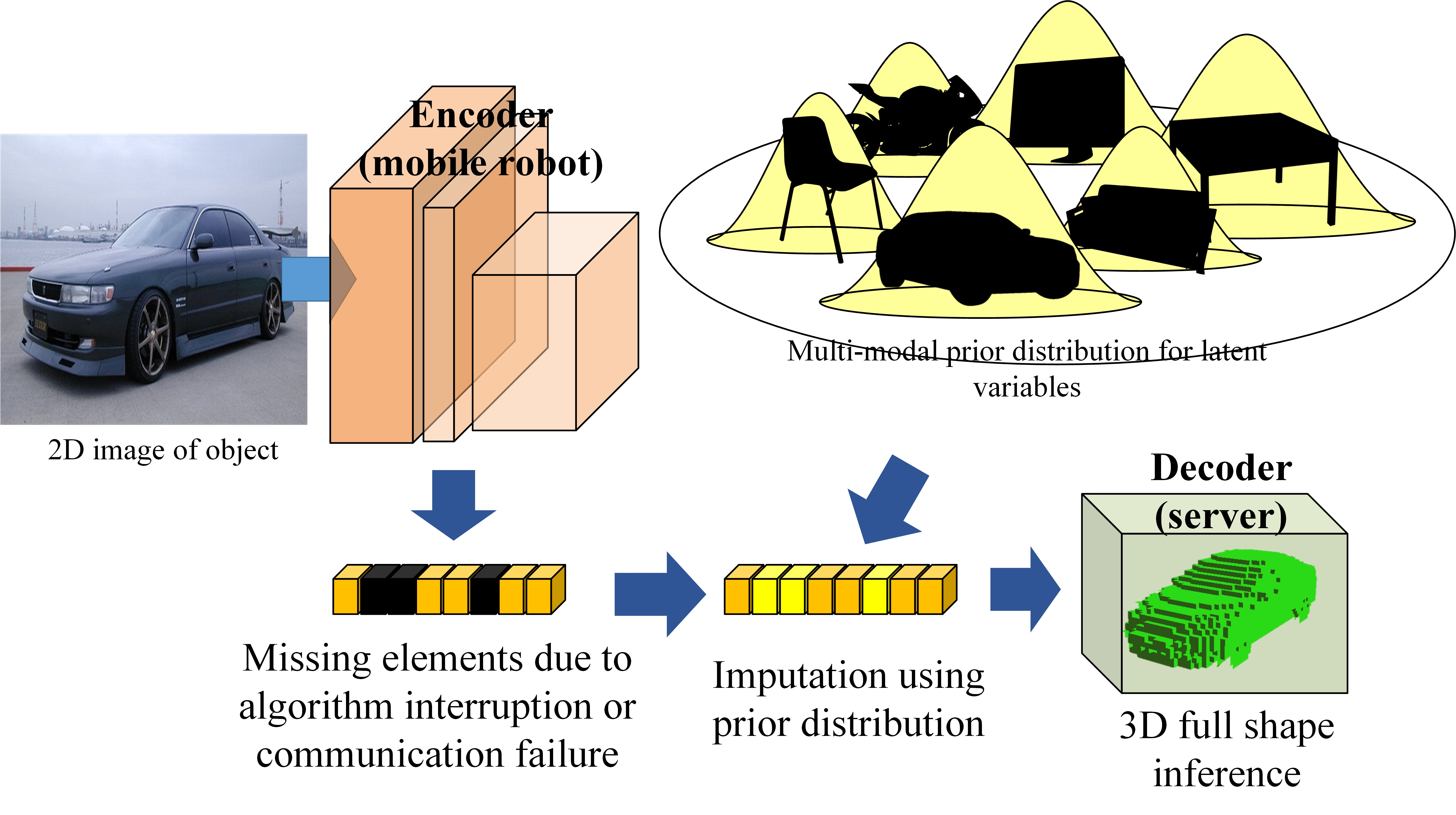}%
		\caption{
			An overview of the proposed method.			
			We train VAE with a multi-modal prior distribution.
			By using the intact elements of the transmitted vector and the prior, we can find the correct modal to perform imputation.
			The supplemented vectors can be subsequently converted to a 3D shape by the decoder.
		}%
		\label{method_overview}%
		\vspace{-15pt}
	\end{figure}%
	\begin{figure*}[t]%
		\centering%
		\includegraphics[scale=0.12]{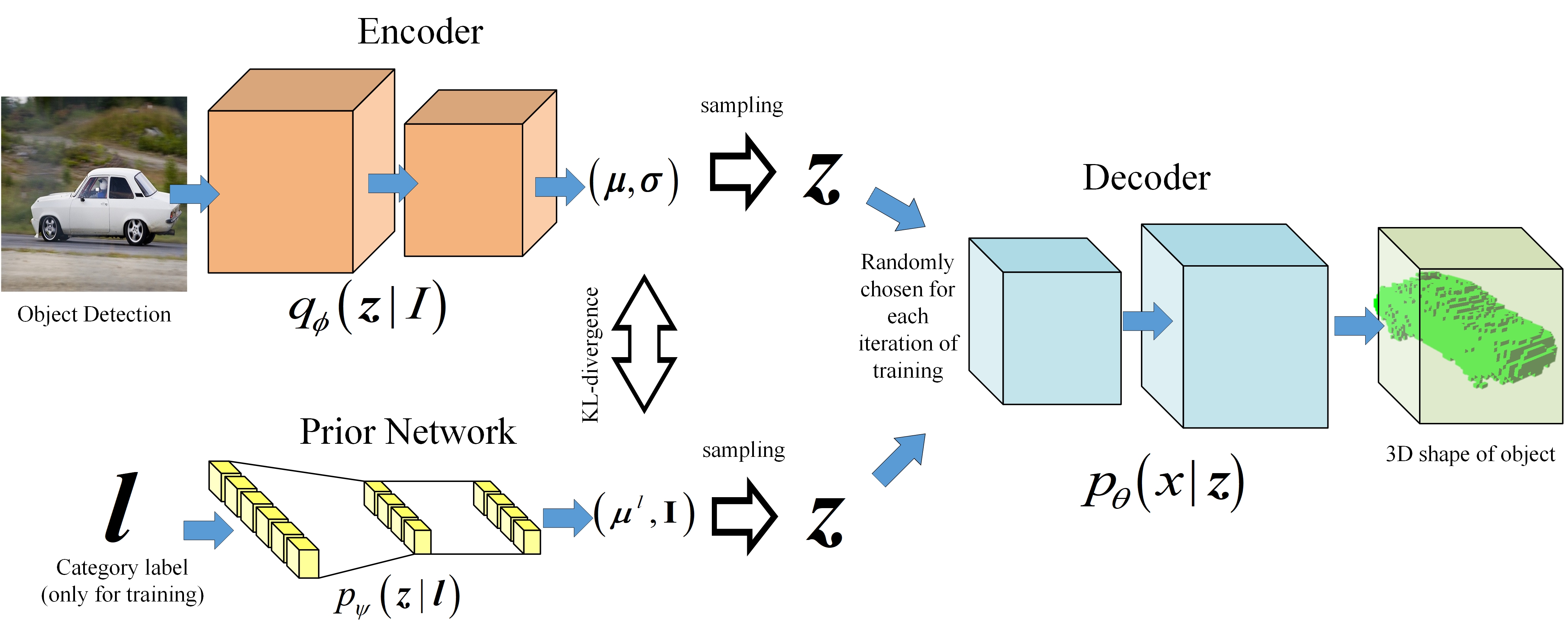}%
		\caption{%
			An overview of the proposed network. During training, the prior network is also trained that represents a multi-modal prior distribution. The encoder can be equipped on a remotely operating robot, while the prior network and 3D decoder are utilized in the base server for a human operator.
			Each dimension of the latent space is assumed to be independent to each other so that target modal of prior distribution can be found by exploiting only a subset of the elements of the latent variable.
			We can, therefore, perform the element-wise imputation of over-compressed vectors.
		}%
		\label{network_overview}%
		\vspace{-10pt}
	\end{figure*}
	
	In the remote human-robot teaming context, it is challenging to support real-time sharing of 2D or 3D perception from a robot in a limited communication environment as the amount of visual or 3D sensory data is significantly larger when compared to that of wave, text, or other low-dimensional signals.
    In this case, most deep network structures can be utilized to obtain compressed latent variables, since a deep network is an inference functions constructed by compositing multiple non-linear functions.
    The network can be divided into two parts at any intermediate layer, and the low-dimensional output of that layer can be used as the compressed latent variables.
	Therefore, we can regard a deep network as an encoder-decoder architecture where latent variables compressed from the 2D or 3D observation by the encoder can be converted to the 3D shape using the decoder~\cite{myIROS2018,myICRA2019,marrnet,image2mesh,3D_GAN, wang2018pixel2mesh, popov2020corenet}.
	The observed objects can be compressed to a low-dimensional latent vector by using the encoder embedded on an on-board computer of a robot.
    With this characteristic, the AE structure can be adopted for data compression and data transmission to address the bottleneck issue in the communication network.
    Rather than transmitting the entire 2D or 3D information, telecommunication can be performed more efficiently in real-time by transmitting only the compressed vectors.
	These vectors can easily be reconstructed to the 3D shape by the decoder on the remote human operator's end.
	
	However, due to the nature of the robot application utilized for real-time in various challenging environment, the algorithm may be interrupted during data encoding, or only a portion of the encoded vector may be transmitted due to a communication failure.
	In these cases AE structure is hard to be used reliably since the trained encoder and decoder stick to the fixed latent space and its dimension.
	Therefore, with the existing approach it is challenging to perform robust reconstruction, and anytime prediction cannot be guaranteed \cite{zilberstein1996using, larsson2017fractalnet}.
	
    
	In this paper, we further address a challenge of having anytime property \cite{larsson2017fractalnet} for 3D reconstruction. 
	To completely perform the reconstruction with only partial elements of the encoded vector, we introduce the missing element imputation approach.
	Our approach considers not only the latent space for the entire training datapoints, but also category-specific distributions for the missing data imputation task.
    We verify our approach on AE structure, since most 3D-3D \cite{brock2016generative,zhang2021pvt,ran2021learning} or 2D-3D networks \cite{marrnet, image2mesh, 3D_GAN, wang2018pixel2mesh, popov2020corenet} can be considered as AE by utilizing intermediate outputs as latent variables.
    In the case of AE (or VAE), we can collect the latent variables obtained from the training data by categorical order; modal for each category can be obtained.
	Therefore, after training, the closest modal to the latent variable whose dimension is partially lost can be found, which denotes the label of the latent vector.
	By sampling the missing elements from that modal, missing data imputation can be performed.
	In other words, we can consider the characteristics of a specific category or instance while performing imputations.
	
	However, the approaches using AE or VAE do not guarantee that the latent space is well-separated by categories.
	It cannot be guaranteed that the modal of the corresponding category can be found with only the remaining elements of the transmitted vector.
	Therefore, we exploit the idea of category-specific multi-modal prior for VAE~ \cite{myIROS2018,myICRA2019,yu2019zero}.
	Each dimension is assumed to be independent in latent space, and each element is trained to be projected onto a category-specific multi-modal distribution,
	i.e., our purpose is to train the network for element-wise category clustering.
	The latent vector is restored from the imputation process by finding the correct modal with partial elements of the incomplete latent variable.
	These restored latent variables can be converted to the fully reconstructed 3D shapes by the decoder.
	We show the training process of the proposed method in Fig.~\ref{method_overview}.
	
	In runtime, our method is proceeded as follows:
	First, imputation for the missing elements is performed by using category-specific multi-modal prior.
	Second, 3D shapes of the object are reconstructed from the retrieved latent variables using the decoder.
	Our method can be applied to robust 3D shape estimation against both the data loss due to unstable networks and the partial discard due to arbitrary compression.
	Evaluated on the ModelNet and Pascal3D datasets for 3D-to-3D and 2D-to-3D cases respectively, our method achieves outstanding performance over autoencoder up to 70\% data loss. 
	
\section{Related work}
For the 3D-3D or 2D-3D estimation, diverse techniques have been studied~\cite{myIROS2018,myICRA2019,marrnet,image2mesh,3D_GAN, brock2016generative,zhang2021pvt,ran2021learning}.
	In this case, the network can be viewed as an AE structure by choosing any intermediate layer as the end of the encoder part; the encoder represents an observed 2D or 3D sensory data into an abstract latent space obtained from the intermediate layer, whereas the decoder estimates the 3D shape from the latent space.
	Here, each pair of encoder and decoder shares an intermediate vector and its latent space.
	In this way, latent variables compressed from the object observation by the encoder can be converted to the 3D shape using the decoder.
    We exploit such a characteristics of the AE structure to adopt it for data compression and data transmission specifically under a harsh network condition.


	Generally, in the AE, the latent space is determined by the distribution of the dataset.
	Intuitively, a sampling-based method in a latent space can be used to perform imputation of the missing element~\cite{qiu2020genomic,camino2019improving,friedjungova2020missing,ma2020midia}.
	The main concern here is that the distribution of the latent space is hardly represented as a closed form, so it is inevitable for the actual imputation approximation to utilize the statistical approaches such as using the average of latent variables.
	In the case of variational autoencoder (VAE), a prior distribution for a latent space can be manually defined during the training time~\cite{VAE}.
	Since the distribution is generally assumed to be isotropic Gaussian, imputation can be performed by sampling from the prior distribution for the omitted elements.
	By using this aspect that a generative model has a tractable prior distribution, many studies of missing data imputation have been conducted in various fields~\cite{mccoy2018variational,shen2020nonlinear,xie2019supervised}.
	In addition, some approaches uses an encoding paradigm that is ranked per order of magnitude that makes the reconstruction with partial data automatic if the connection were to be dropped \cite{marin2020instant}.
	
	Even with the methods above, it still remains challenging to handle discared elements.
	Due to the characteristic of object-oriented features, category- or instance-level characteristics are highly correlated to 3D shape reconstruction.
    However, the previous approaches only consider the entire latent space regardless of the categories, which makes it hard to exploit category-level characteristics.
	Based on this intuition, we utilize a generative model with multi-modal prior, which involves categorical characteristics in the latent space.
	With this category-specific multi-modal prior, missing element imputation can be performed not by considering the redundant parts of the entire latent space, but by the target distribution according to the category.

\section{Approach}

For robots in extreme environment, transmitting 3D perception through wireless communication is challenging.
Since deep networks can drastically abstract the visual understanding by utilizing intermediate outputs as features or latent variables, they can be fully adopted for reducing the amount of data.
However, even if the data can be compressed, in the case of robots used in mines, aviation or deep sea, it is hard to avoid the situation that the communication has been lost at unexpected moments and then reconnected, resulting in a situation where only a portion of the data is transmitted.

To accomplish a robust reconstruction with only partially transmitted data, it is desired to restore the omitted elements of latent variables.
The prior for a latent space can be learned for a generative model, and then missing element imputation can be performed using this prior.
To meet these needs, we propose a method of missing data imputation for 3D shapes by retrieving missing elements from the prior distributions.

\subsection{Prior of AE and VAE for Element Imputation}
For the object representation, let $\boldsymbol{I}$ and $\boldsymbol{x}$ denote the observed 2D or 3D sensory data and its 3D shape, respectively;
let $\boldsymbol{z}$ be the $N$ dimensional latent vector transmitted from the encoder.
Assume that some of the elements of $\boldsymbol{z}$ have been missed due to sudden algorithm interruption.

When the incomplete vector is simply inputted into the decoder, however, it is hard to expect an accurate result as the decoder has been trained for the complete latent space.
In order to approximately retrieve the incomplete latent variable, missing elements can be compensated for by sampling from the latent space.
In AE, however, there is not enough prior information as the AE does not prescribe the prior distribution which reflects categories, but its latent space is simply centered around zero vector.
Even in the case of vanilla VAE, the prior is simply assumed to be isotropic Gaussian,
$p\left(\boldsymbol{z}\right) = N\left(\boldsymbol{z};\boldsymbol{0},\boldsymbol{I}\right)$.
In this case, missing elements is retrieved by sampling from $p\left(\boldsymbol{z}\right)$ for the incomplete latent variable.
Here, the average of the sampled latent variables is closed to the zero vector, as the prior distribution is defined as isotropic.
Then we can approximately perform data imputation for the latent variable with missing elements as the following:
\begin{align}
	\label{imputation_vae}
	\boldsymbol{z}^{\prime} = 
	\begin{cases}
		\boldsymbol{z}^{\prime}_i = 0, &\text{ if } \boldsymbol{z}_i\notin\mathbb{N} \\
		\boldsymbol{z}^{\prime}_i = \boldsymbol{z}_i, &\text{else}
	\end{cases}
\end{align}
where $\boldsymbol{z}$ is the transmitted vector with missing elements; $\boldsymbol{z}^{\prime}$, the retrieved vector by imputation; and $\boldsymbol{z}_i$ is the $i$'th element of the vector $\boldsymbol{z}$.
$\mathbb{N}$ denotes the set of the elements transmitted normally.

In this case, the imputation result only concerns the distribution of the entire latent space, as it is hard to catch the distributions of each datapoint's category.
To achieve the prior knowledge of category and perform robust imputation, we assume that each category-specific modal follows Gaussian as:
\begin{align}
	\label{conditional_prior_AEVE}
	p\left(\boldsymbol{z}|\boldsymbol{l}\right) = N\left(\boldsymbol{z}; \boldsymbol{\mu}_{\boldsymbol{l}}, \boldsymbol{\Sigma}\right)
\end{align}
where $\boldsymbol{\mu}_{\boldsymbol{l}}$ and $\boldsymbol{\Sigma}$ are the parameters of the Gaussian distribution. In our case we assume that $\boldsymbol{\mu}_{\boldsymbol{l}}$ depends on the category label $\boldsymbol{l}$.
Here, we can calculate $\boldsymbol{\mu}_{\boldsymbol{l}}$ by averaging all latent vectors obtained from training data that belong to the category $\boldsymbol{l}$. For simplicity, we assume $\boldsymbol{\Sigma}$ is a constant vector.

\subsection{Category-specific Multi-modal Prior for Element Imputation}
Since we assume a specific prior distribution of the latent variables obtained from the training data, we can approximately have the distributions of 3D shape $\boldsymbol{x}$ as follows:
\begin{align}
	\label{distribution_of_x}
	\nonumber
	p\left(\boldsymbol{x}|\boldsymbol{l}\right) & = \int p_\theta\left(\boldsymbol{x}|\boldsymbol{z}\right) p\left(\boldsymbol{z}|\boldsymbol{l}\right) d\boldsymbol{z} \\
	                & \simeq \frac{1}{N}\sum^{k=N}_{\boldsymbol{z}^k \sim p\left(\boldsymbol{z}|\boldsymbol{l}\right)} p_\theta\left(\boldsymbol{x}|\boldsymbol{z}^k\right)
\end{align}
where $\theta$ denotes the parameter of the decoder. $N$ denotes the number of the sampling points and $\boldsymbol{z}^k$ is the $k$'th sampled datapoint.
However, there are critical limitations that assuming prior distributions of any hidden layers of deep networks including AE and VAE as multi-modal Gaussian prior for multiple categories;
first, there is no guarantee that each cluster follows Gaussian.
Second, each cluster is very close together. This characteristics makes it challenging to find the cluster to which the latent variable belongs.
Consequently, robust imputation and reconstruction are hardly achieved.

Due to these reasons, the category-level shape retrieval becomes challenging.
To achieve the prior knowledge of category or instance which guarantees that each modal follows Gaussian and far from each other, we exploit the VAE with multi-modal prior distribution according to the category label of each object \cite{myIROS2018,myICRA2019,yu2019zero}. This prior can be denoted as:
\begin{align}
	\label{conditional_prior}
	p_\psi\left(\boldsymbol{z}|\boldsymbol{l}\right) = N\left(\boldsymbol{z}; \boldsymbol{\mu}_\psi\left(\boldsymbol{l}\right), \boldsymbol{I}\right),
\end{align}
where $\psi$ is the trainable network parameter of the prior.
The prior distribution is multi-modal prior, and it can be represented as the conditional distribution of the label as in Eq.~\eqref{conditional_prior}.
Different from Eq.~\eqref{conditional_prior_AEVE}, $\boldsymbol{\mu}_\psi\left(\boldsymbol{l}\right)$ is the function of the label $\boldsymbol{l}$.
This function is implemented as a prior network in Fig.~\ref{network_overview}.
The prior network is designed to automatically find the parameters of each modal.
At the very beginning of the training, parameters are initialized randomly.
Those parameters are the outputs of the network according to the categories and they can be updated by training; with KL-divergence loss, each modal in the prior distribution attracts the latent variables from its category, and it also follows the latent variables as well.
With additional restriction loss \cite{myICRA2019,yu2019zero}, each modal is enforced to follow Gaussian, and also moves far from each other in order to be distinguished from each other.
After training, we can simply obtain $\boldsymbol{\mu}$ for each category by inputting $\boldsymbol{l}$ to the trained prior network in advance to the actual runtime.
Then, the target distribution of 3D shape $p\left(\boldsymbol{x}\right)$ can be represented as:
\begin{align}
    \nonumber
    \log p\left(\boldsymbol{x}|\boldsymbol{l}\right)
    \geq
    &-KL
    \left(
    q_\phi
    \left(
        \boldsymbol{z} | I
    \right)
    ||
    p_\psi
    \left(
    \boldsymbol{z} |\boldsymbol{l}
    \right)
    \right)
    \\
    &+
    \mathbb{E}_{\boldsymbol{z}\sim q_\phi}
    \left[
    \log p_\theta
    \left(
    \boldsymbol{x} | \boldsymbol{z}
    \right)
    \right]
    \label{lower_bound}
\end{align}
where $\phi$ denotes the parameter of the encoder.

According to the mean-field theory, we can assume that each element of the latent vector follows independent Gaussian. Therefore, we can choose the closest modal only with partial element of a latent variable and perform imputation as follows:
\begin{align}
	\label{imputation_mmVAE}
	\boldsymbol{z}^{\prime} = 
	\begin{cases}
		\boldsymbol{z}^{\prime}_i = \mu_i^{\prime}, &\text{ if } \boldsymbol{z}_i\notin\mathbb{N} \\
		\boldsymbol{z}^{\prime}_i = \boldsymbol{z}_i, &\text{else}
	\end{cases}
\end{align}
where $\boldsymbol{\mu}^{\prime} = \boldsymbol{\mu}_\psi\left(\boldsymbol{l}^{\prime}\right)$ is the mean of the closest modal to the latent variable $\boldsymbol{z}$.
In the case of VAE, variational likelihood $q_\phi\left(\boldsymbol{z}|\boldsymbol{x}\right)$ approximates the posterior $p\left(\boldsymbol{z}|\boldsymbol{x},\boldsymbol{l}\right)$.
The networks are trained to fit the variational likelihood to the prior distribution as in Eq.~\eqref{lower_bound}, the prior distribution also approximates the posterior to some extent.
Consequently, when the modal $p_\psi\left(\boldsymbol{z}|\boldsymbol{l}\right)$ is chosen correctly, it also means that the conditional posterior $p\left(\boldsymbol{z}|\boldsymbol{x},\boldsymbol{l}\right)$ is also chosen well, which leads to the correct imputation.
Once the latent variable is retrieved properly using the prior, the 3D shape can be estimated using the decoder trained on the latent space.

\subsection{Modal Selection}
The key of retrieving the incomplete vector is to find the prior modal corresponding to the original latent variable.
According to the mean field theorem, each dimension of the latent space can be assumed to be independent.
Therefore, for the incomplete latent variable $\boldsymbol{z}$, optimal label $\boldsymbol{l}^\prime$ corresponding to the original $\boldsymbol{z}$ can be found by comparing the modal of the prior in element-wise manner as follows:
\begin{align}
    \nonumber
	\boldsymbol{l}^\prime &= \argmax_{\boldsymbol{l}} \prod_{\boldsymbol{z}_i\notin\mathbb{N}} p\left(\boldsymbol{z}_i | \boldsymbol{l}_i\right)\\
	&=\argmin_{\boldsymbol{l}} \sum_{\boldsymbol{z}_i\notin\mathbb{N}}|\boldsymbol{z}_i - \boldsymbol{\mu}_i|^2
	\label{optimal_label}
\end{align}
In other words, the category- or instance-level classification is performed only with those elements of latent variables and multi-modal prior where the latent variable is not missing.
Since we assume that each modal of the prior is Gaussian, summations of the element-wise distance are calculated and compared.
In order to make this approach hold, each modal of the prior distribution in the latent space should be separated from each other by a certain distance threshold or more.
To meet this condition, we give an additional constraint between two different labels $\boldsymbol{l}^j$ and $\boldsymbol{l}^k$ while training multi-modal VAE as in~\cite{myIROS2018,myICRA2019,yu2019zero}:
\begin{align}
\label{mean_distance_thresh}
	|\boldsymbol{\mu}_\psi\left(\boldsymbol{l}^j\right)_i - \boldsymbol{\mu}_\psi\left(\boldsymbol{l}^k\right)_i| > \sigma , \text{ } \forall i,j,k,\text{ }j\neq k
\end{align}
From Eq.~\eqref{mean_distance_thresh}, each dimension of the latent space follows an independent multi-modal distribution, and each modal becomes distinguishable according to the label.
Consequently, target modal can be found using only some non-missing elements of the latent variable, and element-wise imputation can be achieved from this selected modal.

\subsection{Dropout for Element Pruning}
Our method is for anytime robust reconstruction with only partial elements of the datapoint.
In the same vein, other learning schemes such as weight pruning or channel masking can be applied \cite{han2015deep, yan2015hd, hinton2015distilling, kim2018nestednet}.
The purpose of those methods are different since they does not perform imputation and are applied to speech recognition or classification.
But the context is similar in that they use partial elements or partial networks.
Therefore, in our method, element pruning or element masking can be adopted during training so that we perform element imputation and reconstruction from insufficient elements as well. 
Such a element-wise pruning or masking can be simply implemented using dropout, and randomized element-wise pruning can be obtained by setting the dropout rate at random.
Therefore, the decoder is trained to perform reconstruction even when some elements of the vector are pruned, so that more robust anytime reconstruction algorithm can be achieved.

\begin{table*}[t]%
	\caption{Classification results of incomplete latent variables by using Euclidian distance to the multi-modal prior}%
\vspace{-10pt}
	\label{distance_eval}%
	\begin{center}%
		\begin{tabular}{c|c c c c c c c}
			\hline
			&& discard rate & 0\% (base) & 30\% & 50\% & 70\% & 90\% \\
			\hline
			\multirow{6}{*}{ModelNet40~\cite{modelnet}}
			&& AE & 0.6088 & 0.5498 & 0.4792 & 0.3668 & 0.1630 \\
			&& VAE & 0.6331 & 0.5453 & 0.4716 & 0.3280 & 0.1420 \\
			&& AE\_dropout & 0.6091 & 0.5693 & 0.5191 & 0.4240 & 0.2157 \\
			&& VAE\_dropout & 0.6076 & 0.5555 & 0.5026 & 0.4050 & 0.2044 \\
			\cline{2-8}
			&& ours & \textbf{0.8020} & \textbf{0.7531} & \textbf{0.6773} & \textbf{0.5308} & \textbf{0.2352} \\
			&& ours\_dropout & 0.7354 & 0.6830 & 0.6096 & 0.4848 & 0.1983 \\
			\hline
			\multirow{6}{*}{Pascal3D+~\cite{pascal3D}}
			&& AE & 0.9775 & 0.9620 & 0.9311 & 0.8431 & 0.5161 \\
			&& VAE & 0.9699 & 0.9535 & 0.9250 & 0.8317 & 0.5125 \\
			&& AE\_dropout & 0.9759 & 0.9728 & 0.9655 & 0.9383 & \textbf{0.7294} \\
			&& VAE\_dropout & 0.9766 & 0.9708 & 0.9639 & 0.9317 & 0.6807 \\
			\cline{2-8}
			&& ours & \textbf{0.9788} & \textbf{0.9766} & \textbf{0.9699} & \textbf{0.9397} & 0.6933 \\
			&& ours\_dropout & 0.9784 & 0.9747 & 0.9670 & 0.9215 & 0.6479 \\
			\hline
		\end{tabular}%
	\end{center}%
\end{table*}%
\begin{table*}[t]
	\caption{Precision and Recall evaluation with various discard rate}\vspace{-10pt}%
	\label{precision_recall_table}%
	\begin{center}
		\begin{tabular}{c |c cc c cc c cc c cc c cc}
			\hline
			\hline
			\multirow{2}{*}{dataset}  &  & \multicolumn{2}{c}{0\% (base)} && \multicolumn{2}{c}{30\%} && \multicolumn{2}{c}{50\%} && \multicolumn{2}{c}{70\%} && \multicolumn{2}{c}{90\%} \\
			&& precision & recall && precision & recall && precision & recall && precision & recall && precision & recall\\
			\hline
			& AE     & 0.7888 & 0.7951 && 0.7051 & 0.6516 && 0.5922 & 0.4792 && 0.4564 & 0.3205 && 0.2741 & 0.1670 \\
			& VAE    & \textbf{0.8331} & \textbf{0.8610} && 0.6214 & 0.5774 && 0.4716 & 0.4099 && 0.3366 & 0.2781 && 0.2165 & 0.1745\\
			ModelNet40& AE\_dr & 0.7477 & 0.4445 && 0.6523 & 0.2898 && 0.5619 & 0.2073 && 0.4511 & 0.1486 && \textbf{0.3370} & 0.1001\\
            ~\cite{modelnet}& VAE\_dr& 0.7409 & 0.2471 && 0.6186 & 0.1550 && 0.5228 & 0.1172 && 0.4331 & 0.0945 && \textbf{0.3370} & 0.0659\\
            \cline{2-16}
			& ours   & 0.8048 & 0.8527 && 0.7468 & \textbf{0.7476} && 0.6723 & \textbf{0.6038} && \textbf{0.5322} & \textbf{0.3880} && 0.3048 & \textbf{0.1544}\\
			&ours\_dr& 0.8287 & 0.5294 && \textbf{0.7668} & 0.3577 && \textbf{0.6827} & 0.2160 && 0.5315 & 0.1344 && 0.3175 & 0.0722\\
			\hline
			& AE     & 0.8105 & 0.8177 && 0.7630 & 0.7788 && 0.7008 & 0.7237 && 0.5956 & 0.6220 && 0.3799 & 0.4008\\
			& VAE    & 0.8065 & 0.8221 && 0.7468 & 0.7624 && 0.6739 & 0.6888 && 0.5544 & 0.5625 && 0.3561 & 0.3549\\
			Pascal3D+& AE\_dr & 0.8088 & 0.8028 && 0.7796 & 0.7812 && 0.7322 & 0.7388 && 0.6500 & 0.6607 && 0.4437 & \textbf{0.5062}\\
            \cite{pascal3D}& VAE\_dr& 0.8063 & 0.8329 && 0.7729 & 0.8080 && 0.7198 & 0.7596 && 0.6356 & 0.6978 && 0.4577 & 0.4869\\
			\cline{2-16}
			& ours   & 0.8046 & 0.8208 && 0.7935 & 0.8151 && 0.7654 & 0.7909 && \textbf{0.6965} & \textbf{0.7246} && \textbf{0.4619} & 0.4795\\
			&ours\_dr& \textbf{0.8127} & \textbf{0.8234} && \textbf{0.8003} & \textbf{0.8153} && \textbf{0.7726} & \textbf{0.7927} && 0.6930 & 0.7177 && 0.4416 & 0.4591\\
			\hline
		\end{tabular}%
	\end{center}
\end{table*}
\begin{table*}[t]
	\caption{Effect of imputation for Precision and Recall}\vspace{-8pt}%
	\label{effect_of_imputation}%
	\begin{center}
		\begin{tabular}{c |c |cc cc cc | cc cc cc }
			\hline
			\hline
			\multirow{3}{*}{dataset}  &  & \multicolumn{6}{c}{50\% missing elements}  & \multicolumn{6}{c}{70\% missing elements}  \\
			&  & \multicolumn{2}{c}{before imputation} & \multicolumn{2}{c}{after imputation} & \multicolumn{2}{c}{improved (\%)} & \multicolumn{2}{c}{before imputation} & \multicolumn{2}{c}{after imputation} &\multicolumn{2}{c}{improved (\%)} \\
			&& pre & rec & pre & rec &pre&rec & pre & rec & pre & rec & pre&rec\\
			\hline
			& AE     & 0.6173 & 0.3900 & 0.5922 & 0.4792 &\textcolor{red}{-4.24}& \textcolor{blue}{18.61} & 0.4980 & 0.1841 & 0.4564 & 0.3205 &\textcolor{red}{-9.11}&\textcolor{blue}{42.56}\\
			& VAE    & 0.6581 & 0.3665 & 0.4716 & 0.4099 & \textcolor{red}{-39.54}&\textcolor{blue}{10.59} & 0.5271 & 0.1413 & 0.3366 & 0.2781 &\textcolor{red}{-56.60}&\textcolor{blue}{49.19}\\
			ModelNet40& AE\_dr & 0.5416 & 0.1650 & 0.5619 & 0.2037 &\textcolor{blue}{3.61}&\textcolor{blue}{19.00} & 0.4325 & 0.0901 & 0.4511 & 0.1486 &\textcolor{blue}{4.12}&\textcolor{blue}{39.37}\\
            ~\cite{modelnet}& VAE\_dr& 0.5137 & 0.0675 & 0.5228 & 0.1172 &\textcolor{blue}{1.74}&\textcolor{blue}{42.41} & 0.4259 & 0.0335 & 0.4331 & 0.0945 &\textcolor{blue}{1.66}&\textcolor{blue}{64.55}\\
			& ours   & 0.7062 & 0.4152 & 0.6723 & 0.6038 &\textcolor{red}{-5.04}&\textcolor{blue}{31.24} & 0.5330 & 0.1118 & 0.5322 & 0.3880 &\textcolor{red}{-0.15}&\textcolor{blue}{71.19}\\
			&ours\_dr& 0.6703 & 0.0540 & 0.6827 & 0.2160 &\textcolor{blue}{1.82}&\textcolor{blue}{75.00} & 0.4556 & 0.0193 & 0.5315 & 0.1344 &\textcolor{blue}{14.28}&\textcolor{blue}{85.64}\\
			\hline
			& AE     & 0.6521 & 0.5782 & 0.7008 & 0.7237 &\textcolor{blue}{6.95}&\textcolor{blue}{20.11} & 0.5054 & 0.3752 & 0.5956 & 0.6220 &\textcolor{blue}{15.14}&\textcolor{blue}{39.68}\\
			& VAE    & 0.6841 & 0.6624 & 0.6739 & 0.6888 &\textcolor{red}{-1.51}&\textcolor{blue}{3.83} & 0.5391 & 0.4601 & 0.5544 & 0.5625 &\textcolor{blue}{2.76}&\textcolor{blue}{18.20}\\
			Pascal3D+& AE\_dr & 0.8021 & 0.5359 & 0.7322 & 0.7398 &\textcolor{red}{-9.55}&\textcolor{blue}{27.83} & 0.7379 & 0.1892 & 0.6500 & 0.6609 &\textcolor{red}{-13.52}&\textcolor{blue}{71.37}\\
            \cite{pascal3D}& VAE\_dr& 0.7931 & 0.6973 & 0.7198 & 0.7596 &\textcolor{red}{-10.18}&\textcolor{blue}{8.20} & 0.7660 & 0.3783 & 0.6356 & 0.6978 &\textcolor{red}{-20.52}&\textcolor{blue}{45.79}\\
			& ours   & 0.7553 & 0.7091 & 0.7654 & 0.7909 &\textcolor{blue}{1.32}&\textcolor{blue}{10.34} & 0.6593 & 0.4561 & 0.6965 & 0.7246 &\textcolor{blue}{5.34}&\textcolor{blue}{37.05}\\
			&ours\_dr& 0.7646 & 0.7507 & 0.7726 & 0.7927 &\textcolor{blue}{1.04}&\textcolor{blue}{5.30} & 0.6614 & 0.5541 & 0.6930 & 0.7177 &\textcolor{blue}{4.56}&\textcolor{blue}{22.80}\\
			\hline
		\end{tabular}%
	\end{center}
\end{table*}

\subsection{Decoder and Prior Distribution}
After training is completely converged, we can find the category-specific modal $p_\psi\left(\boldsymbol{z}|\boldsymbol{l}\right)$ of the incomplete latent variable and let the latent variable be supplemented.
Subsequently, the robust 3D reconstruction can then be achieved by the decoder.
However, since it is challenging for the variational likelihood $q_\phi\left(\boldsymbol{z}|\boldsymbol{x}\right)$ to accurately approximate the prior $p\left(\boldsymbol{z}|\boldsymbol{x},\boldsymbol{l}\right)$ in practice, adapting the decoder to the prior distribution as well can flexibly cope with the latent variables under the imputation process.
Therefore, we replace the expectation term in Eq.~\eqref{lower_bound} with the following:
\begin{align}
\label{modified_expectation_term}
	\mathbb{E}_{\boldsymbol{z}\sim q_\phi\left(\boldsymbol{z}|\boldsymbol{x}\right)}
    \left[
    \log p_\theta
    \left(
    \boldsymbol{x} | \boldsymbol{z}
    \right)
    \right]
    +
    \mathbb{E}_{\boldsymbol{z}\sim p_\psi\left(\boldsymbol{z}|\boldsymbol{l}\right)}
    \left[
    \log p_\theta
    \left(
    \boldsymbol{x} | \boldsymbol{z}
    \right)
    \right]
\end{align}
By Eq.~\eqref{modified_expectation_term}, the decoder also estimates the 3D shape from the latent variable sampled from the prior distribution according to the label.
With this modification, when the incomplete latent variable is supplemented by replacing the omitted element with the variables from the prior, we can obtain more robust 3D reconstruction results.
In the actual training phase, those two expectation terms are not trained at the same time and randomly selected per one training iteration.

\section{Experiments}
In order to verify the proposed method, we use the ModelNet40 dataset~\cite{modelnet} for 3D object observation, transmission and 3D reconstruction and Pascal3D dataset~\cite{pascal3D} for object detection in 2D image, transmission and 3D estimation.
Modelnet contains 40 classes and about 300 instances for each category, and Pascal3D includes 10 classes and 10 instances for each category.
We set latent dimension to $64$ for Pascal3D and ModelNet40.
While transmitting the latent variable, some elements can be rejected at various rates due to unexpected interruption, or for artibrary over-compression rate.
Therefore, in this experiment, the rejection ratios (or probability) of elements are set to 30, 50, 70, and 90\%.
For the 3D shape information, we convert CAD model into $64^3$ voxel grids with binary variables.
Since there are also the images of multi-object scenes in Pascal3D dataset, we crop the images to obtain single-object images using bounding boxes.
The size of the train and test images is set to $448\times 448$.

We analyze the 3D reconstruction results using the decoder, after performing classification for modal selection, and missing element imputation.
The case of using AE and vanilla VAE are also evaluated for comparison.
We follow Eq.~\eqref{imputation_vae} for VAE when performing missing element imputation of latent variables.
In the case of AE, since there is no assumption of the latent space, we simply assume that the prior distribution is Gaussian similar to VAE.
As mentioned in Section III, in the case of AE and VAE, category-specific modals are obtained in advance to the runtime by averaging latent variables obtained from training dataset according to each category.
For our method, category-specific modals are also estimated in advance to the runtime, by using the prior network and one-hot category labels.

\subsection{Classification} 
The proposed method aims to achieve robust 3D shape reconstruction from the incomplete latent variable whose elements are omitted.
To handle this issue, it is important to find the modal corresponding to the label of the object with only exploiting the elements that remain from the original vector.
In other words, the possibility of performing correct 3D reconstruction increases when label classification (or modal selection) using Eq.~\eqref{optimal_label} is successfully performed.
We evaluate the label classification accuracy by finding the nearest modal with the remaining elements of the latent variable.

Table~\ref{distance_eval} shows the results of classifications for two datasets.
Classifications are performed using Eq.~\eqref{optimal_label}.
In the test time dropout is also applied to the missing elements in the latent space.
For ModelNet40, our method shows higher accuracy rate compared to AE- and VAE-based methods. 
In our method, dimensions are assumed to be independent to each other and each element follows a one-dimensional multi-modal prior, so the classifications tasks are performed relatively well even in the cases where most of the elements of the latent variables have been lost.
Interestingly, dropout generally increases the performance for the case of AE and VAE as if we adopt dropout for the classification task, but does not bring certain effects to our method. 

When a half of the dimensions are lost, the accuracies reached 65\% or more for ModelNet40.
Even the classification is conducted only with 10\% of the elements, the method achieved almost 23.5\% accuracy.
This indicates that even when the latent variable fails to accurately follow the class-wise multi-modal distribution independently for each dimension, the exact modal according to the label of the object can be estimated with only a few dimensions of the latent vector.
However, for Pascal3D+, the proposed method performs better but does not show high performance gap compared to the case of ModelNet40.
We believe that Pascal3D+ only has 10 classes so that it is a little bit easier to perform classification compared to the case of ModelNet40 that has 40 classes. Also, Pascal3D+ has high-resolution RGB images as input which is more easier to extract feature-rich information compared to the case of 3D models in low-resolution.

Compared to the 3D reconstruction, the classification task showed a higher success rate as the task follows a regression for a much simpler multinoulli distribution rather than the multi-dimensional binary estimation for complex 3D grids.

\begin{figure*}[t]%
	\centering%
	\includegraphics[scale=0.31]{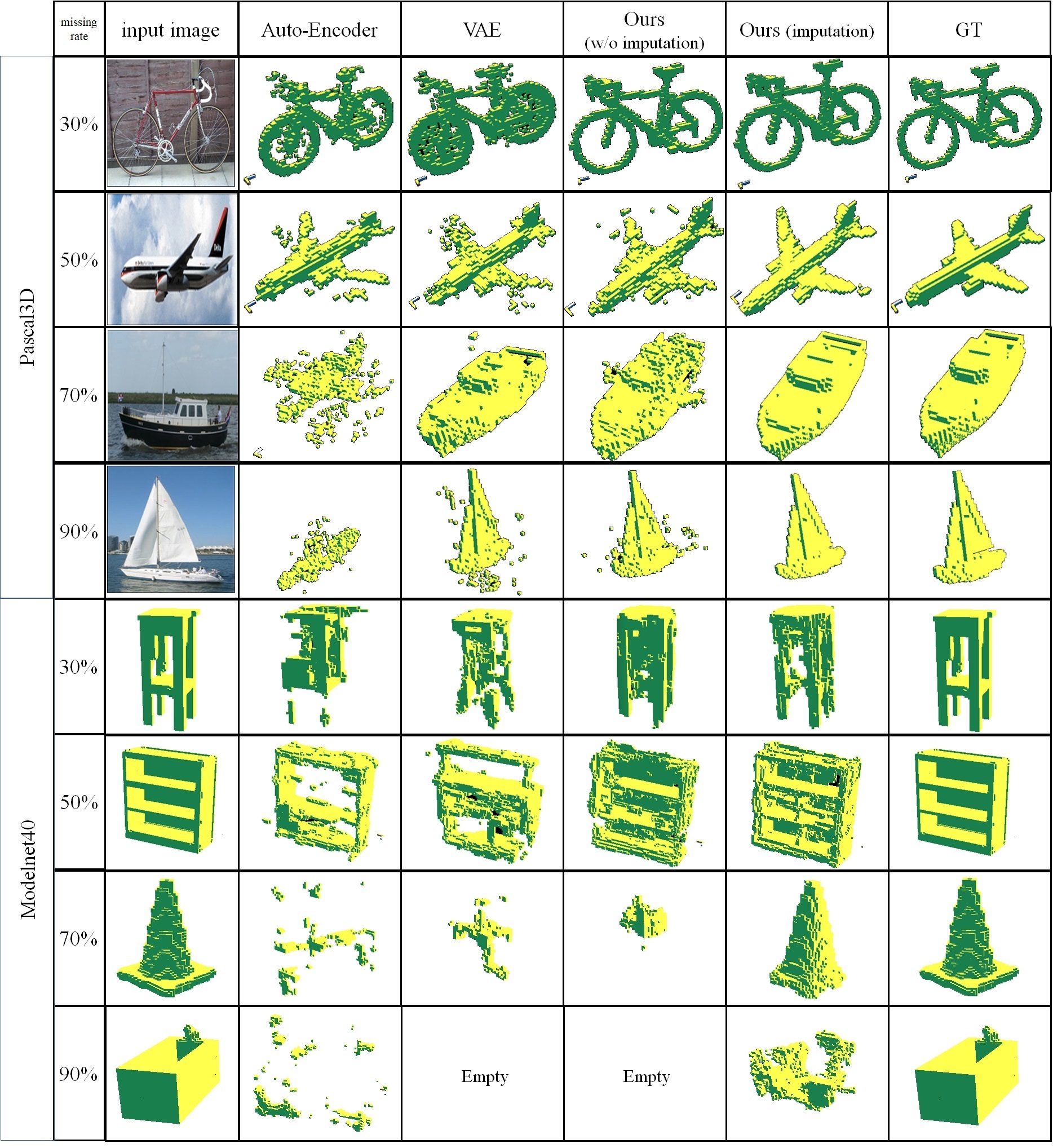}%
	\caption{%
		Examples of 3D shape reconstruction.%
	}%
	\label{shape_eval}%
\end{figure*}

\subsection{Reconstruction}
\subsubsection{Reconstruction after Imputation}
We represent the quantitative results of 3D shape reconstruction in Table.~\ref{precision_recall_table}.
$dr$ denotes dropout.
Similar to the classification task, the precision-recall results are obtained for various missing rates, 30, 50, 70, and 90\%.
In Table.~\ref{precision_recall_table}, we display the precision-recall results with voxel occupancy threshold $\lambda=0.5$.
Since our proposed method retrieves discarded elements based on the rest elements and prior distribution, the method achieves highest recall rate while preserving its precision rate high enough for all cases.
We also found that dropout for element pruning increases the performance evaluated on Pascal3D+, rather decreases in the case of ModelNet40.
We believe that this masking and pruning are usually proposed for 2D image classification task \cite{kim2018nestednet, he2017channel}, and we hardly tell that those methods can enhance the anytime reconstruction algorithm for high-dimensional 3D input.

In addition to quantitative results, 3D shape estimation examples are shown in Fig.~\ref{shape_eval}.
In the case of 30 and 50\% discard rate, the results indicate that the proposed method achieves robust reconstruction results.
We found that the result shows blurred or empty reconstruction when the discard rate exceeds 70\%, similar to the case of the precision-recall evaluation.
Since ModelNet dataset is more challenging than Pascal3D, this trend is particularly noticeable on ModelNet.
In consideration of this, we manually select the showcase examples where the proposed method almost completely reconstruct the 3D shape despite of the extremely high loss rate of the latent variable.
\subsubsection{Effect of Imputation}
We also report the effect of imputation.
Our method assumes cateogory-level multi-modal prior and performs imputation using this prior. Therefore, for the comparison we conduct na\"ive imputation under the assumption that the latent space follows unimodal Gaussian and we do not have multi-modal prior.
Regardless of the methods, the mean of Gaussian was similar to the zero vector.
We show the evaluation results in Table.~\ref{effect_of_imputation}.
For clarity, results for 50\% and 70\% are displayed.
We represent the increased performance as blue, and decreased one as red.
Recall rates are highly improved for all cases.
In ModelNet40, all methods except AE and VAE shows better reconstruction performance which achieves high improvement of recall with maintaining the same precision as the case of before imputation.
In Pascal3D+, our methods show significant improvement of the all recall score.
We also take advantages in the aspect of the memory efficiency and the computational time. Since our method do not require any prior 3D models (in our case $64^3$ boolean) for each categories but use latent variables (64-dim float) for the prior distribution, it only uses $0.78$\% memory compared to the case of using prior 3D models directly.
To find the closest modal, any sorting algorithm can be used after calculating Euclidean distance; it only takes $O\left(N\log N\right)$. In addition, the number of the category or instance is a constant (in our case 10 or 40) and it only takes under a few millisecond.

\section{Conclusion}
In the context of human-robot teaming in harsh environments or  low-bandwidth communication networks, as in disaster-response or military domains, real-time object observation and transmission may be interrupted or failed so that only partial elements of compressed data can be transferred.
To support robust real-time human-robot teaming even under unstable environments, we propose an anytime reconstruction method by considering the category-specific multi-modal distribution.
Although Autoencoder (AE) and Variational Autoencoder (VAE) have been exploited as key structures to compress and decode data, imputing lost elements in the aspect of category or instance is challenging due to the simplicity of their prior distribution.
To achieve a category-level imputation and complete 3D shape reconstruction, we exploit the idea of multi-modal prior distribution for a latent space.
Different from the vanilla VAE, each modal in the proposed approach is determined automatically while training, and contains information of specific category.
Using this prior distribution, we determine the modal of latent variables merely with the transmitted elements in the latent space.
By imputing discarded elements with sampled variables from the chosen modal, we can robustly achieve latent vector retrieval and 3D shape reconstruction.
\clearpage


\bibliographystyle{IEEEtran}
\bibliography{bibmy}

\end{document}